%% file: root.tex
\documentclass[letterpaper, 10 pt, journal, twoside]{IEEEtran}

\usepackage{url}

\usepackage{times}
\usepackage{epsfig}
\usepackage{graphicx}
\usepackage{amsmath}
\usepackage{amssymb}
\usepackage{cite}

\usepackage{xcolor}
\usepackage{caption}
\usepackage{rotating}
\usepackage{verbatim}
\usepackage{xcolor,colortbl}
\usepackage{etoolbox}
\usepackage[normalem]{ulem}
\usepackage{booktabs}
\usepackage{multirow}
\usepackage{lipsum}
\usepackage{stmaryrd}
\usepackage{stackengine}
\usepackage{makecell}
\usepackage{pifont}
\usepackage{cancel}
\usepackage{adjustbox}
\usepackage{dblfloatfix}
\usepackage{marvosym}

\usepackage{hyperref}

\hypersetup{
  colorlinks=true,
  linkcolor=blue,
  citecolor=blue,
  urlcolor=black
}

\begin{document}

\markboth{IEEE ROBOTICS AND AUTOMATION LETTERS. PREPRINT VERSION. ACCEPTED January, 2026}{Jin \MakeLowercase{\textit{et al.}}: OccTENS}

\title{\LARGE \bf
OccTENS: 3D \underline{Occ}upancy World Model via \underline{Te}mporal \underline{N}ext-\underline{S}cale Prediction
}

\author{Bu Jin\(^{\ast}\), Songen Gu\(^{\ast}\), Xiaotao Hu\(^{\ast}\), Yupeng Zheng, Xiaoyang Guo, Qian Zhang, Xiaoxiao Long, Wei Yin$^{\text{\Letter}}$
\thanks{Manuscript received July 10, 2025; Revised November 3, 2025; Accepted January 12, 2026.
This paper was recommended for publication by the Editor, Prof. Hyungpil Moon, upon evaluation of the reviewers' comments.}
\thanks{Bu Jin is with The Hong Kong University of Science and Technology.
Songen Gu and Yupeng Zheng are with the University of Chinese Academy of Sciences.
Xiaoyang Guo, Qian Zhang, and Wei Yin are with Horizon Robotics.
Xiaotao Hu is with The Hong Kong University of Science and Technology.
Xiaoxiao Long is with the Nanjing University.}
\thanks{Bu Jin\(^{\ast}\), Songen Gu\(^{\ast}\), and Xiaotao Hu\(^{\ast}\) contribute equally. Wei Yin$^{\text{\Letter}}$ is the corresponding author. Email: yvanwy@outlook.com}
\thanks{Digital Object Identifier (DOI): see top of this page.}
}

\maketitle

%%%%%%%%%%%%%%%%%%%%%%%%%%%%%%%%%%%%%%%%%%%%%%%%%%%%%%%%%%%%%%%%%%%%%%%%%%%%%%%%
\input{sections/0_abs}

\input{sections/1_intro}

\input{sections/2_relatedwork}

\input{sections/3_method}

\input{sections/4_exp}

\input{sections/5_conclusion}




% \section*{ACKNOWLEDGMENT}

% The preferred spelling of the word ÒacknowledgmentÓ in America is without an ÒeÓ after the ÒgÓ. Avoid the stilted expression, ÒOne of us (R. B. G.) thanks . . .Ó  Instead, try ÒR. B. G. thanksÓ. Put sponsor acknowledgments in the unnumbered footnote on the first page.

% %%%%%%%%%%%%%%%%%%%%%%%%%%%%%%%%%%%%%%%%%%%%%%%%%%%%%%%%%%%%%%%%%%%%%%%%%%%%%%%%

% References are important to the reader; therefore, each citation must be complete and correct. If at all possible, references should be commonly available publications.

\bibliographystyle{IEEEtran}
\bibliography{main}

\end{document}

%% file: sections/0_abs.tex
\begin{abstract}

% emphasize open world instead of LLM

% Recent advancements in autonomous driving have yielded promising results, yet challenges remain in handling long-tail distributions and out-of-distribution scenarios. World models, which simulate and understand the environment by learning a comprehensive representation of the external world, offer a promising solution to these challenges. Among them, occupancy world models have gained attention for their ability to predict regions of space occupied by objects, enabling spatial understanding essential for navigation and decision-making. However, existing approaches often overlook the scale of objects, treating all entities equally without considering their real-world size. This simplification can lead to inaccurate predictions, particularly in environments where objects of varying scales coexist.

In this paper, we propose OccTENS, a generative occupancy world model that enables controllable, high-fidelity long-term occupancy generation while maintaining computational efficiency.
Different from visual generation, the occupancy world model must capture the fine-grained 3D geometry and dynamic evolution of the 3D scenes, posing great challenges for the generative models.
Recent approaches based on autoregression (AR) have demonstrated the potential to predict vehicle movement and future occupancy scenes simultaneously from historical observations, but they typically suffer from \textbf{inefficiency}, \textbf{temporal degradation} in long-term generation and \textbf{lack of controllability}. To holistically address these issues, we reformulate the occupancy world model as a temporal next-scale prediction (TENS) task, which decomposes the temporal sequence modeling problem into the modeling of spatial scale-by-scale generation and temporal scene-by-scene prediction. With a \textbf{TensFormer}, OccTENS can effectively manage the temporal causality and spatial relationships of occupancy sequences in a flexible and scalable way.
To enhance the pose controllability, we further propose a holistic pose aggregation strategy, which features a unified sequence modeling for occupancy and ego-motion.
Experiments show that OccTENS outperforms the state-of-the-art method with both higher occupancy quality and faster inference time.
% \textcolor{red}{Add specific data.}

\end{abstract}


%% file: sections/1_intro.tex
\section{Introduction}

% \textcolor{red}{Background.} 
\IEEEPARstart{R}{ecent} years have seen significant advancements in the development of autonomous driving (AD) systems. While existing AD methods \cite{hu2023planning, jiang2023vad, hu2022st, huang2023differentiable, yang2024unipad} have demonstrated excellent results across a range of driving scenarios, there are still challenges when dealing with long-tail distributions or out-of-distribution situations. A promising direction to address these challenges is the world model \cite{wang2024drivedreamer, gao2024vista, gao2023magicdrive, hu2023gaia, wang2024driving, yang2023bevcontrol, swerdlow2024street}, which simulates and comprehends the surrounding environment by learning a comprehensive representation of the external world. 

Occupancy world model, as a specialized type of world model, has gained significant attention for its expressiveness of the 3D geometry.  
Several occupancy world models \cite{zheng2023occworld, wei2024occllama, wang2024occsora, xu2025occ, gu2024dome, li2024uniscene, agro2024uno, diehl2025dio} have been developed in recent years. 
For example, some works \cite{zheng2023occworld, wei2024occllama, xu2025occ} have attempted to tailor GPT-like architecture to generate occupancy in an autoregressive manner.
However, when generating long-time sequences, they exhibit \textbf{temporal degradation} and \textbf{inefficiency} issues due to the increasing accumulation of tokens over time. Moreover, they lack the ability to control generation results based on a specified camera pose or trajectory, which limits the model’s predictive capacity for varying viewpoints and its reasoning ability regarding the agent’s location and orientation.

\begin{figure}[!t]
    \centering
    \includegraphics[width=\linewidth]{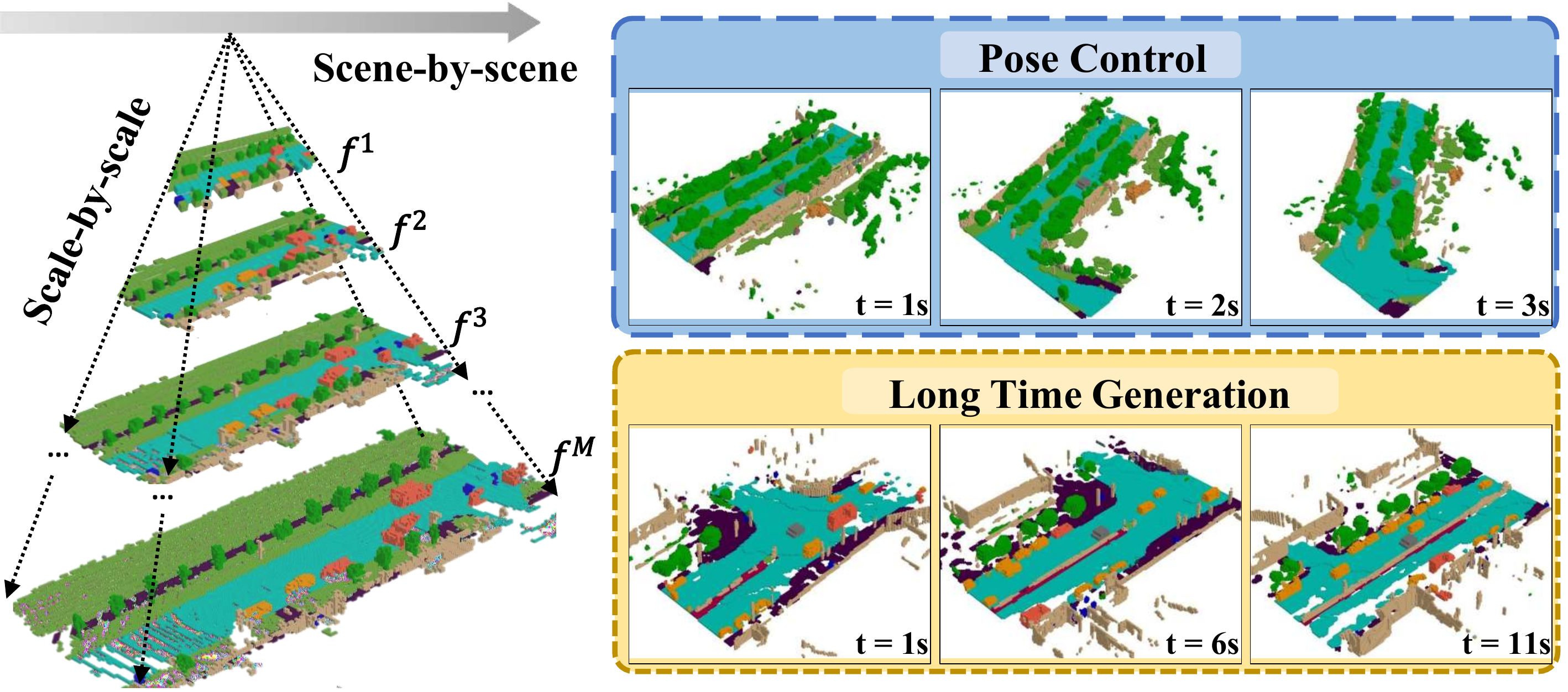}
    \caption{We propose OccTENS, a coarse-to-fine occupancy world model that enables controllable, high-fidelity long-time occupancy generation while maintaining computational efficiency.}
    \vspace{-6mm}
    \label{fig:teaser}
\end{figure}

To address these issues, we introduce \textbf{OccTENS}, a novel autoregressive occupancy world model designed to generate controllable, high-fidelity long-term occupancy scenes while maintaining computational efficiency. Our key innovation lies in reformulating the occupancy world model as a temporal next-scale prediction (TENS) task, which decomposes the temporal sequence modeling problem to the modeling of spatial scale-by-scale generation and temporal scene-by-scene prediction. 
This decomposition enables OccTENS with both computational efficiency and high-fidelity long-term occupancy scene generation through a principled balance of parallelizable spatial refinement and sequential temporal reasoning.
Further, we propose a multi-modal camera pose aggregation module tailored for auto-regressive models, which facilitates pose controllability and motion planning simultaneously.

While the temporal next-scale prediction in OccTENS is inspired by next-scale prediction in VAR\cite{tian2024visual}, it is \textbf{non-trivial} at all from VAR to OccTENS:  
(1) \textit{Lack of temporal modeling}. While the original VAR was primarily designed for a class-to-image task, we explore its extension to temporal sequences, allowing it to capture dynamic, time-varying patterns beyond purely spatial next-scale prediction.
(2) \textit{Temporal degradation}. The multi-scale design naturally requires longer token sequences, which can overload attention mechanisms and result in temporal degradation during long-term generation. Instead, OccTENS decouples the frame regression from scale regression, enabling a more effective modeling of both temporal causality across different timestamps and spatial relationships across various scales.
(3) \textit{Lack of multi-modal interaction}. A world model should understand the trajectory or motion of the ego vehicle to plan a trajectory or control future occupancy generation.
(4) \textit{Representation discrepancy}. 3D occupancy involves a more complex topological structure than 2D images, such as the geometric continuity or sparse nature, highlighting the superiority of OccTENS as a critical advancement for the occupancy world model.

Our contributions are summarized as follows:
\begin{itemize}
    \item We introduce \textbf{OccTENS}, a coarse-to-fine occupancy world model that enables controllable, high-fidelity long-term occupancy generation while maintaining computational efficiency. 
    \item We reformulate the occupancy world model as a temporal next-scale prediction (TENS) task, which decomposes the temporal sequence modeling problem to the modeling of spatial scale-by-scale generation and temporal scene-by-scene prediction. With a \textbf{TensFormer}, OccTENS can effectively manage the temporal causality and spatial relationships of occupancy sequences.
    \item We propose a holistic camera pose aggregation strategy tailored for auto-regressive models. With a unified sequence modeling for occupancy and camera pose, OccTENS facilitates pose controllability (controlling occupancy with a given trajectory) and motion planning (planning a trajectory for the AD vehicle) simultaneously.
    \item Extensive experiments demonstrate the superior performance and efficiency of OccTENS on the occupancy prediction task and motion planning task.
\end{itemize}

%% file: sections/2_relatedwork.tex
\section{Related Work}
\vspace{-1mm}

\subsection{Occupancy Prediction and Generation}
Recently, the 3D occupancy prediction task \cite{wei2023surroundocc, tong2023scene, cao2022monoscene, huang2023tri, ma2024cotr, li2023fb, wang2024panoocc, zheng2024monoocc, pan2024renderocc, huang2024selfocc} has gained increasing attention, as it enhances scene understanding with both geometric and semantic information.
MonoScene \cite{cao2022monoscene} is the pioneer in vision-based occupancy prediction. Subsequent research has advanced occupancy prediction accuracy and reduced computational overhead by improving scene representation \cite{huang2023tri, ma2024cotr}, enhancing view transformation modules \cite{li2023fb}, integrating object detection \cite{wang2024panoocc}, introducing multi-view 2D information \cite{pan2024renderocc, huang2024selfocc}, and employing privileged learning \cite{zheng2024monoocc}.
Recent studies \cite{zheng2023occworld, wei2024occllama, gu2024dome, wang2024occsora, xu2025occ, li2024uniscene} have adopted occupancy as the representation of world models in autonomous driving, aiming at generating future occupancy based on historical observation. Occworld \cite{zheng2023occworld} is the earliest approach for the occupancy world model. OccLlama \cite{wei2024occllama} and Occllm \cite{xu2025occ} enhance the generation performance by incorporating large language models. OccSora \cite{wang2024occsora} and Dome \cite{gu2024dome} taming diffusion models to generate future occupancy based on trajectory input. UnO \cite{agro2024uno} and DIO \cite{diehl2025dio}leverage implicit formulations to model 3D scenes as continuous occupancy fields, enabling perception and forecasting at an infinite resolution. DIO builds upon this concept by introducing a decomposable 4D world model that jointly represents occupancy and scene flow over time. Uniscene \cite{li2024uniscene} further introduces a unified scene prediction module including occupancy, video, and lidar prediction, requiring the BEV layouts as input.  However, the demand for the ground truth of future trajectories or BEV layouts makes it impossible for the downstream motion planning task.  Moreover, although these works are well-designed architectures, they suffer from either low-fidelity or low-efficiency problems. It remains challenging to \textit{simultaneously} achieve high-fidelity occupancy estimation and fast inference time.

\subsection{World Model}
World model \cite{ha2018world} can predict the consequences of various actions, which is crucial for autonomous driving. Traditional models emphasize visual prediction \cite{hu2023gaia, zhao2024drivedreamer, su2024text2street, gao2024vista, wang2023drivedreamer, lu2023wovogen, wang2024driving, zheng2024genad, jiang2024dive, deng2024autoregressive, li2024autoregressive}, potentially overlooking the essential 3D information needed for AD vehicles. Some approaches attempt to forecast point clouds using unannotated LiDAR scans \cite{zhang2023learning, zyrianov2024lidardm}, but these methods neglect semantic information and are not suitable for vision-based or fusion-based autonomous driving. Occupancy world models \cite{zheng2023occworld, wei2024occllama, wang2024occsora, diehl2025dio, agro2024uno} create a world model in 3D occupancy space, providing a more comprehensive understanding of the evolvement of 3D scenes, which is the main focus of our work.

\begin{figure*}[!t]
    \centering
    \includegraphics[width=1\textwidth]{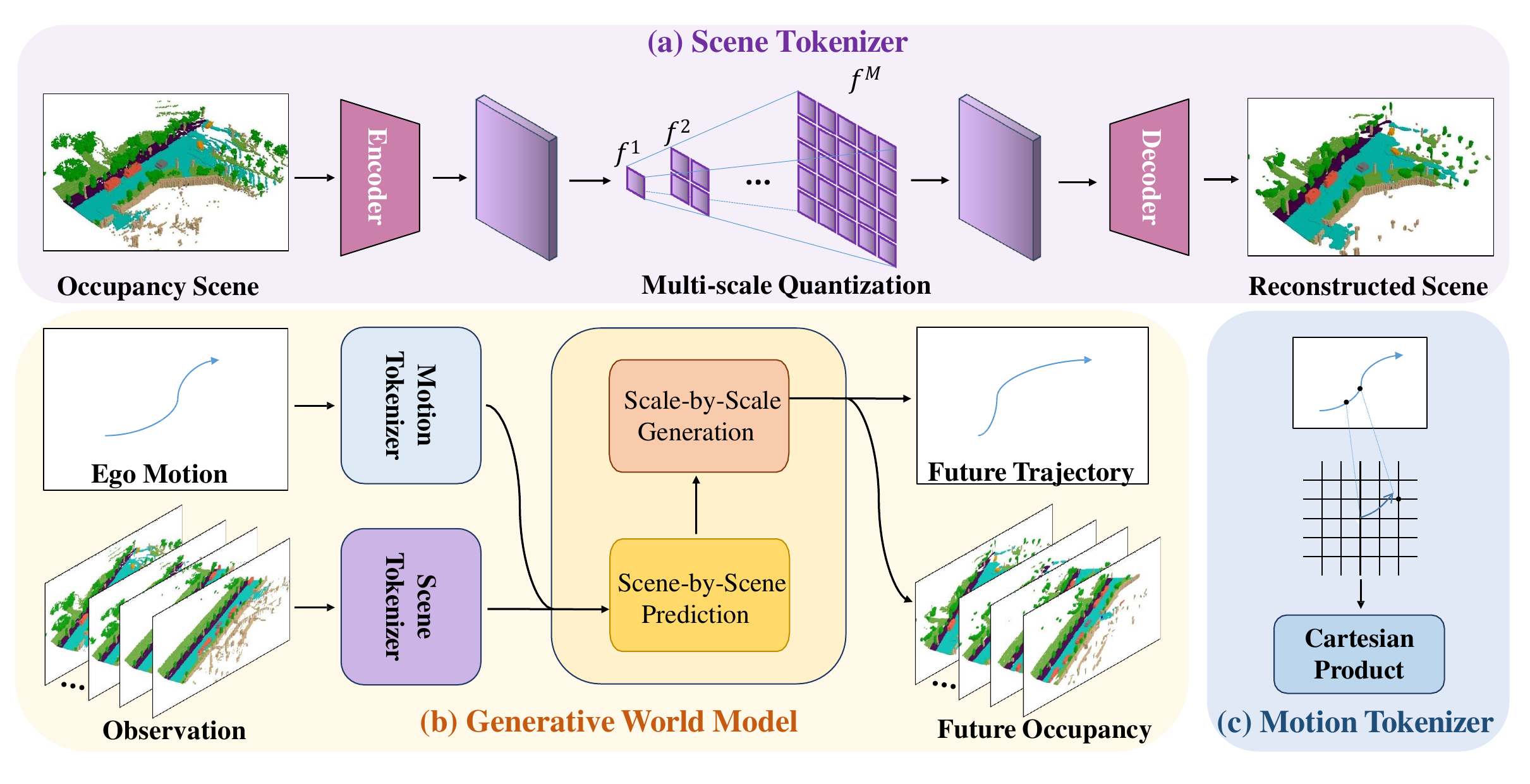}
    \caption{\textbf{Overview of OccTENS}. OccTENS consists of two components: two tokenizers (a and c) that encode 3D occupancy and ego motion into discrete tokens, and a generative world model (b) using temporal next-scale prediction for future 3D scene forecasting.}
    \vspace{-4mm}
    \label{fig:main}
\end{figure*}

% \subsection{Visual Auto-regressive Generation}
% Visual Autoregressive (AR) Generation refers to utilizing autoregressive methods to generate images \cite{esser2021taming, razavi2019generating, lee2022autoregressive, yu2021vector, yu2022scaling} or videos \cite{yan2021videogpt, kalchbrenner2017video, yu2023language}. Generally, the AR models employ a raster-scan paradigm, which encodes and flattens 2D images into 1D token sequences. 
% Recently, VAR \cite{tian2024visual} proposes to utilize next-scale prediction in visual auto-regressive modeling, which effectively handles the inefficiency and spatial degradation of next-token prediction. However, it still remains unclear whether the next-scale prediction is suitable for occupancy prediction. Moreover, the modeling process of VAR does not take account of the temporal dimension, which still requires a consideration for how to avoid the temporal degradation of this modeling approach, which is the main focus of our work.

% \subsection{Multi-scale Neural Architectures}
% Multi-scale spirits have attracted researchers' attention for a long time and are widely applied in computer vision and 2D generation. For example, Multiscale Vision Transformers \cite{fan2021multiscale} introduces a multiscale pyramid of features to transformer layers, which produces promising results in image and video recognition. 

% \subsection{Discussion}
% 对比现在的方法和VAR的方案
% 对比pyramid和occsora
% 对比occ预测里的multi-scale
% 展示结果和速度对比

%% file: sections/3_method.tex
\section{Method}

In this work, we propose \textbf{OccTENS}, a novel occupancy world model designed to comprehend historical observations and forecast the future 3D scenarios. As illustrated in Fig.~\ref{fig:main}, our proposed OccTENS consists of two components: a robust tokenizer that encodes 3D occupancy and ego motion into discrete tokens (see Sec.~\ref{subsec:tokenization}), and a generative world model using next-scale prediction for future scene forecasting and motion planning (see Sec.~\ref{subsec:world_model}).

\subsection{Tokenizer}
\label{subsec:tokenization}
The goal of the tokenizer is to model the 3D occupancy scene and the ego motion as discrete tokens.

\subsubsection{Scene Tokenizer}
The goal of the scene tokenizer is to model the 3D occupancy scene as discrete tokens. To achieve this, a common practice for scene tokenizer is to employ a quantized autoencoder like VQVAE \cite{zheng2023occworld, wei2024occllama}, which quantizes the occupancy feature map with discrete feature vectors.
%TODO: 这里要讨论一下，因为var里没写 next-scale prediction 里的 multi-scale quantizer 对 tokenization 的影响，只写了是要适配于 next-scale prediction
However, unlike natural language sentences with an inherent left-to-right ordering, the occupancy feature maps are inter-dependent, resulting in the bidirectional correlations of the quantized token sequence. This contradicts the unidirectional dependency assumption of autoregressive models, where each token can only depend on its prefix, as illustrated in \cite{tian2024visual}. Thus, we propose a multi-scale tokenizer specifically designed for next-scale prediction.

% \textbf{Occupancy Encoder.} 
Firstly, we employ an occupancy encoder to encode the occupancy scene into a feature map.
Given a scene $\mathbf{S}$, we convert it to a BEV representation with a series of 2D convolution layers, resulting in a latent feature $\mathbf{F} \in \mathbb{R}^{H \times W \times C}$.
% We employ an embedding layer to embed the 3D occupancy scene into a latent space. Then we convert the 3D scene to a BEV representation by merging the height dimension with the channel dimension. 
% \textbf{Multi-scale Quantizer.} 
We then utilize a quantizer to tokenize the latent feature $\mathbf{F}$ into multi-scale discrete tokens. 
Previous approaches \cite{zheng2023occworld, wei2024occllama} attempt to transform the feature into a collection of codebook entries through vector quantization, where each entry is responsible for a small area.
However, operating tokenization solely on local information may result in the loss of global context. We utilize the multi-scale quantizer \cite{tian2024visual} to discretize the occupancy feature $\mathbf{F} \in \mathbb{R}^{H \times W \times C}$ to $M$ multi-scale discrete token maps: $\mathbf{F}=(\mathbf{f}^1, \mathbf{f}^2, ..., \mathbf{f}^M)$.

The occupancy decoder takes the multi-scale quantized tokens as input and outputs the reconstructed 3D occupancy scenes. 
% Firstly, we convert the multi-scale quantized tokens $\mathbf{F}_q=(\mathbf{f}_q^1, \mathbf{f}_q^2, ..., \mathbf{f}_q^M)$ to the BEV feature map $\mathbf{\hat{F}}$. We upsample the features of each scale to the original resolution and then pass them as input to the convolution layers $\phi_{1,...,M}$, which is the same as those in Eq.~\ref{eq:quantization_upsampling}. The interpolated results are accumulated to get the reconstructed bev feature map. This process can be formulated as:
% \begin{equation}
%     \mathbf{\hat{F}} = \sum_{m=1}^M\phi_m(\mathcal{I}(\mathbf{f}_q^m, (x_d, y_d)))
% \end{equation}
To reconstruct the 3D occupancy scene from the quantized tokens $\mathbf{F}$, we utilize another series of convolution layers to upsample the BEV feature map to the initial resolution, and then split the height dimension from the channel dimension, resulting in the reconstructed scene $\mathbf{\hat{S}}$.

\subsubsection{Motion Tokenizer}
The motion tokenizer is utilized to discretize the pose of the ego vehicle to better integrate it into our sequence prediction model. 
We utilize the position $x,y$ and orientation $\theta$ relative to the previous frame to represent the motion of the vehicle. We discard the information in the z-axis because the vehicle's speed in the z-axis is nearly zero most of the time. We apply a vanilla uniform quantization of the motion information, resulting in $V_x$, $V_y$ and $V_{\theta}$ tokens in the vocabulary. Then we map the relative motion with three discrete tokens to a motion token $\mathbf{P}$ by Cartesian product:
\begin{equation}
    \mathbf{P} = \mathcal{E}(x + y \times V_x + \theta \times V_x \times V_y),
\end{equation}
where $\mathcal{E}$ is an embedding layer.
The discretized poses are then embedded into the sequence prediction model.
% where they are treated as an additional condition for the occupancy prediction. This allows the model to understand how the environment evolves as the vehicle moves, leading to more accurate and consistent predictions, particularly in complex or dynamic environments.

\subsection{Generative World Model}
\label{subsec:world_model}
In this section, we introduce our generative world model via temporal next-token prediction, as shown in Fig.~\ref{fig:main}. 
% As \textbf{input}, we assume the tokenized BEV features $\mathbf{F} = \{\mathbf{F}_1, ..., \mathbf{F}_{T-1}\}$ and ego motion $\mathbf{P} = \{ \mathbf{P}_1, ..., \mathbf{P}_{T-1}  \}$ with previous $T-1$ frames. The \textbf{target output} of the world model is the occupancy scene $\mathbf{S}_T$ and motion $\mathbf{P}_T$ at the $T$-th frame. 
% More frames can be obtained by shifting and repeating the generation process.

\subsubsection{Rethinking Autoregressive Models for Occupancy Generation}
Some previous occupancy world models \cite{wei2024occllama, xu2025occ} utilize a vanilla next-token autoregressive modeling on occupancy prediction. Considering the feature map with resolution $(n \times n)$, the likelihood of the sequence $x=\{x_1, x_2, ..., x_{n \times n}\}$ can be decomposed to the product of $n \times n$ conditional probabilities:
\begin{equation}
p(x) =\prod_{i=1}^{n \times n} p\left(x_t \mid x_1, x_2, \ldots, x_{i-1}\right)
\end{equation}

However, such a vanilla modeling paradigm introduces several issues like inefficiency over longer autoregressive steps.
A naive solution to address the inefficiency issue is to utilize next-scale prediction \cite{tian2024visual}, which has demonstrated successful practice in class-to-image generation. However, the multi-scale nature makes it require longer token sequences. For example, assuming the scales are even distributed, given a feature map with $(n \times n)$ resolution and $M$ scales, the total sequence number is:
\begin{equation}
    n^2\sum_{m=1}^{M}(\frac{m}{M})^2 \approx o(Mn^2)
\end{equation}
When we adapt next-scale prediction for temporal modeling, the increasing token number can overload attention mechanisms, distributing their focus across expanding frame numbers. This results in progressive fragmentation of long-range dependencies, undermining global temporal coherence for long-term generation. 

Moreover, the dependencies associated with scale also necessitate a comprehensive understanding of scale interaction over time. When generating a low-scale token, the model primarily needs to consider the previous low-scale information, as it encapsulates the global structural context like scene layouts that persists coherently over time. In contrast, higher-scale tokens, which capture finer-grained details such as textures, demand a global interaction between their own scale and the lower scales across frames. Thus, a trivial adaptation approach for next-scale prediction to temporal next-scale prediction is never sufficient.

To address the issues, OccTENS introduces a novel architecture, named \textbf{TENSFormer}, by decoupling the next-scale prediction within a single occupancy scene generation from the per-frame prediction across the occupancy sequences. This decoupling enables OccTENS to more effectively manage both temporal causality across different timestamps and spatial relationships across various scales, resulting in a more flexible and scalable framework for temporal occupancy modeling.

\subsection{Architecture of TENSFormer}
\begin{figure}[!t]
    \centering
    \includegraphics[width=0.5\textwidth]{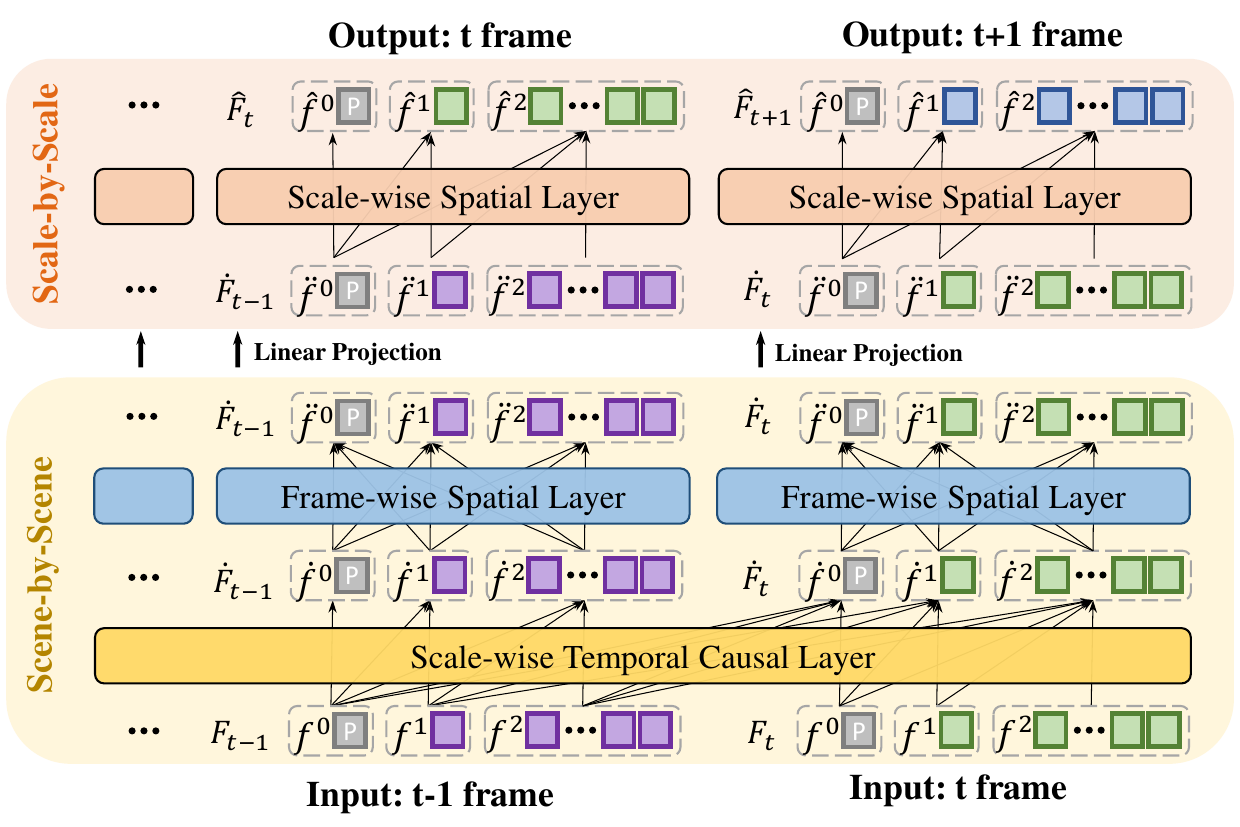}
    % \vspace{-5mm}
    \caption{\textbf{Temporal Next-scale Prediction.} The proposed TENSFormer decomposes the sequential occupancy generation into two distinct components: a scene-by-scene prediction and a scale-by-scale generation. The $\dot{F}$, $\ddot{F}$, $\hat{F}$ denote intermediate representations after the scale-wise temporal causal layer, after the frame-wise spatial layer and predicted logits for the next frame, respectively.}
    \vspace{-4mm}
    \label{fig:model_detail}
\end{figure}

The proposed TENSFormer decomposes the sequential occupancy generation into two distinct components: a scene-by-scene prediction and a scale-by-scale generation, as detailed in Fig.~\ref{fig:model_detail}. The scene-by-scene prediction is designed to model the dependencies between scales across consecutive frames, ensuring that temporal dynamics are accurately represented and integrated. Meanwhile, the scale-by-scale generation focuses on capturing the spatial composition of the occupancy scene within a single frame. To simultaneously facilitate pose control (controlling occupancy with given trajectory) and motion planning (planning a trajectory for the AD vehicle), we further introduce a multi-modal camera pose aggregation module tailored for auto-regressive generative models.

\begin{figure*}[!t]
    \centering
    \includegraphics[width=0.90\textwidth]{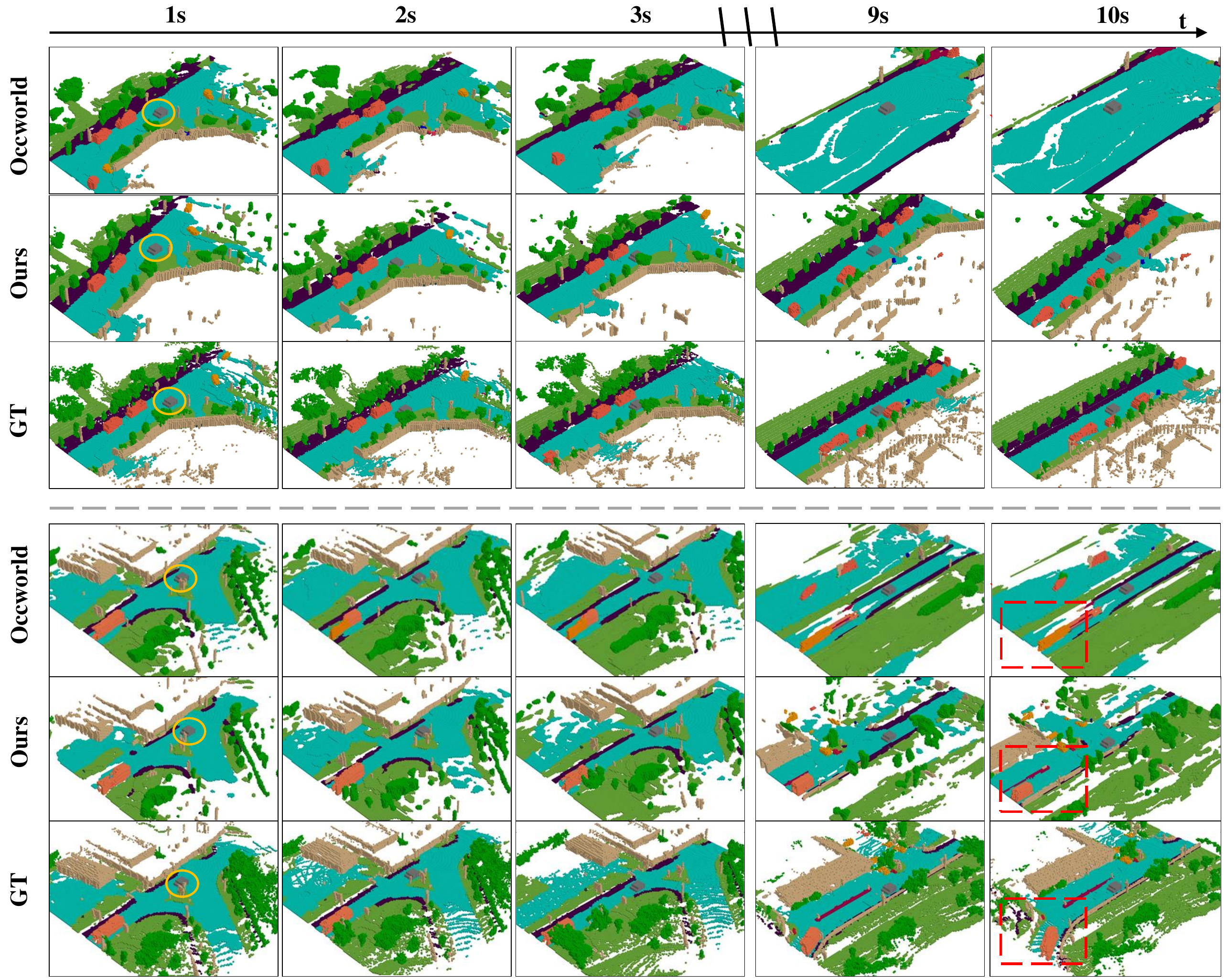}
    \caption{\textbf{Qualitative results for long-term generation.} We compare OccTENS with OccWorld \cite{zheng2023occworld} in generating long sequences. OccWorld exhibits \textcolor{red}{repetition} artifacts. In contrast, OccTENS produces more diverse and realistic occupancy scenes. We mark the ego vehicle with an \textcolor{orange}{orange circle} in the first column.}
    \vspace{-3mm}
    \label{fig:long_generation}
\end{figure*}

\subsubsection{Temporal Scene-by-scene Prediction}
The scene-by-scene prediction module aims to capture the temporal dependencies in occupancy sequences. Occupancy frames are naturally causal sequences, with each frame depending on its preceding frames. Therefore, we utilize a causal attention mechanism to ensure that each frame can only attend to its preceding frames. Given occupancy tokens $\{\mathbf{F}_1, ..., \mathbf{F}_{T-1}\}$ with $T$ frame, the process can be formulated as:
\begin{equation}
p(\mathbf{F}_1, ..., \mathbf{F}_{T})=\prod_{t=1}^{T} p(\mathbf{F}_{t} \mid \mathbf{F}_0, ..., \mathbf{F}_{t-1})
\end{equation}
Note that we add three 1-D sine-cosine embeddings to indicate the position information, scale information, and time information respectively, contributing to the spatial and temporal modeling. 
However, we find that this frame-wise causal attention may result in structure collapse when generating occupancy tokens with large scales, which we attribute to an inherent ambiguity in separating temporal dynamics from spatial dependencies, as the frame-wise causal attention conflates inter-frame causality with intra-frame bidirectional dependency.
Thus we further decompose the frame-wise causal attention into two parts: scale-wise temporal causal attention and frame-wise spatial attention. For scale-wise causal attention, the tokens $\mathbf{f}_{t}^{m}$ of $m$-th scale in $t$-th frame can only attend to its prefix and the tokens in the same scale: $\{\mathbf{F}_1, \mathbf{F}_2, \ldots, \mathbf{F}_{t-1}, \mathbf{f}_{t}^1, \mathbf{f}_{t}^2, \ldots, \mathbf{f}_{t}^{m-1} \}$.
For frame-wise spatial attention, we utilize a full attention to model the scale interaction within each frame. 
In this way, the causal attention and spatial attention are fully decoupled in scene-by-scene prediction, making it possible to synchronously model the temporal dependency for each occupancy scale. This enhances the training efficiency and allows the progressive training strategy \cite{tian2024visual} for effective scale modeling, where hierarchical scale-specific dependencies are incrementally optimized.
% In this process, the attention scores are computed within a scene. We utilize a full-attention mask to ensure that each token $\mathbf{\Dot{f}}_{t}^{m}$ has a global view at the whole scene $\{ \mathbf{\Dot{f}}_{t}^1, \mathbf{\Dot{f}}_{t}^2, \ldots, \mathbf{\Dot{f}}_{t}^{M} \}$. After intra-frame attention blocks, the token of $t$-th frame at $m$-th scale as $\mathbf{\Ddot{f}}_{t}^{m}$.

\subsubsection{Spatial Scale-by-scale Generation}
The scale-by-scale generation module focuses on generating the occupancy given the historical features. In this period, we inherit the outputs of the previous scene-by-scene prediction module as the \textbf{guidance features} to navigate spatial feature generation.  
When generating the $m$-th scale at $t$-th frame, the autoregressive likelihood is formulated as:
\begin{equation}
% p(x)=\prod_{m=0}^{M} p(\mathbf{\hat{f}}_{t+1}^m \mid \mathbf{\Ddot{F}}_t, \mathbf{\hat{f}}_{t+1}^0, \mathbf{\hat{f}}_{t+1}^1, \ldots, \mathbf{\hat{f}}_{t+1}^{m-1})
p(\mathbf{F'}_{t-1}, \mathbf{\hat{f}}_{t}^1, \ldots, \mathbf{\hat{f}}_{t}^{M})=\prod_{m=1}^{M} p(\mathbf{\hat{f}}_{t}^m \mid \mathbf{F'}_t, \mathbf{\hat{f}}_{t}^1, \ldots, \mathbf{\hat{f}}_{t}^{m-1})
\end{equation}
where the $\mathbf{F'}_{t-1}$ is the guidance features and $\mathbf{\hat{f}}_{t}^m$ denotes the $k$-th scale tokens at $t$-th frame. Note that we utilize a block-wise causal attention mask to ensure that each token $\mathbf{\hat{f}}_{t}^{m}$ can only attend to its prefix $\{ \mathbf{F}_{t-1}, \mathbf{\hat{f}}_{t}^1, \ldots, \mathbf{\hat{f}}_{t}^{m-1} \}$.

\subsubsection{Multi-modal Camera Pose Aggregation}
A key challenge for diffusion-based methods \cite{wang2024occsora, gu2024dome} is that they can not conduct the motion planning task because they require the ground-truth future trajectory as a fixed conditioning input throughout diffusion sampling. At the same time, the existing auto-regressive solutions \cite{wei2024occllama, zheng2023occworld, xu2025occ} suffer from controlling the occupancy generation with given camera pose (or trajectory). In OccTENS, we introduce a multi-modal camera pose aggregation module tailored for auto-regressive generative models, which enables motion planning and pose control simultaneously.

For auto-regressive models, we would add a BOS token before occupancy tokens, indicating the beginning of an occupancy frame sequence. 
To better integrate the ego-vehicle motion into our prediction paradigm, we treat the motion token as $0$-th scale token $\mathbf{f}_0$ and splice it before the multi-scale occupancy tokens, resulting in $M+1$ tokens $\{ \mathbf{f}_{t}^0, \mathbf{f}_{t}^{1}, ..., \mathbf{f}_{t}^{M} \}$ for each frame. Thus, the autoregressive likelihood is reformulated as
\begin{equation}
% p(x)=\prod_{m=0}^{M} p(\mathbf{\hat{f}}_{t+1}^m \mid \mathbf{\Ddot{F}}_t, \mathbf{\hat{f}}_{t+1}^0, \mathbf{\hat{f}}_{t+1}^1, \ldots, \mathbf{\hat{f}}_{t+1}^{m-1})
p(\mathbf{F'}_{t-1}, \mathbf{\hat{f}}_{t}^0, \mathbf{\hat{f}}_{t}^1, \ldots, \mathbf{\hat{f}}_{t}^{M})=\prod_{m=0}^{M} p(\mathbf{\hat{f}}_{t}^m \mid \mathbf{F'}_t, \mathbf{\hat{f}}_{t}^0, \mathbf{\hat{f}}_{t}^1, \ldots, \mathbf{\hat{f}}_{t}^{m-1})
\end{equation}

\subsection{Loss Function}
When training the scene tokenizer, we utilize cross-entropy loss and lovasz-softmax loss \cite{berman2018lovasz}. To enhance the global occupancy reconstruction performance, we also utilize geoscal loss and semscal loss illustrated in \cite{cao2022monoscene}, which optimize the class-wise derivable precision, recall and specificity for semantics and geometry. In general, our loss function is defined as:
$
    \mathcal{L} = {\lambda}_1 \mathcal{L}_{ce} + {\lambda}_2 \mathcal{L}_{lovasz} + {\lambda}_3 \mathcal{L}_{geoscal} + {\lambda}_4 \mathcal{L}_{semscal},
$
where the factors ${\lambda}_{1,2,3,4}$ are used to balance the losses.

When training the world model, we utilize cross-entropy loss for the generation of occupancy tokens and pose tokens, defined as:
$
    \mathcal{L} = {\beta}_1 \mathcal{L}_{occ} + {\beta}_2 \mathcal{L}_{pose}
$.

%% file: sections/4_exp.tex
\begin{table*}[!t]
\setlength{\tabcolsep}{0.02\linewidth}
\centering
\renewcommand\arraystretch{1.1}
% \resizebox{\linewidth}{!}{
\begin{tabular}{l|cc|cccc|cccc}
\toprule
\multirow{2}{*}{Method} & \multirow{2}{*}{Dataset} & \multirow{2}{*}{Input} &  \multicolumn{4}{c|}{MIOU(\%)$\uparrow$}  & \multicolumn{4}{c}{IOU(\%)$\uparrow$}    \\
  &  &  & 1s & 2s & 3s & \cellcolor[HTML]{E6E6E6}Avg. & 1s & 2s & 3s & \cellcolor[HTML]{E6E6E6}Avg. \\ 
\midrule
% \multicolumn{9}{c}{\cellcolor[HTML]{FFFFFF}Camera Input} \\
% \midrule
OccWorld-F \cite{zheng2023occworld} & nuScenes & Cam & 8.03  & 6.91   &  3.54  &  \cellcolor[HTML]{E6E6E6}6.16     &     23.62   &  18.13  &  15.22  &  \cellcolor[HTML]{E6E6E6}18.99    \\
OccLLaMA-F \cite{wei2024occllama} & nuScenes & Cam &  10.34  &  8.66  &  6.98  &   \cellcolor[HTML]{E6E6E6}8.66         &  25.81  &  23.19  &   19.97 &   \cellcolor[HTML]{E6E6E6}22.99  \\
\textbf{OccTENS-F (Ours)} & nuScenes & Cam &  \textbf{17.17}  &  \textbf{10.38}  &  \textbf{7.82}  &   \cellcolor[HTML]{E6E6E6}\textbf{11.79}         &  \textbf{27.60}  &  \textbf{25.14}  &  \textbf{20.33} &   \cellcolor[HTML]{E6E6E6}\textbf{24.35}  \\
% \cline{1-9}
\midrule
% \multicolumn{9}{c}{\cellcolor[HTML]{FFFFFF}Occupancy Input} \\
% \cline{1-9}
% \midrule
OccWorld-O \cite{zheng2023occworld} & nuScenes & Occ &  25.78  &  15.14  &  10.51  &  \cellcolor[HTML]{E6E6E6}17.14        &  34.63  &  25.07  &  20.18  &  \cellcolor[HTML]{E6E6E6}26.63            \\
OccLLaMA-O \cite{wei2024occllama} & nuScenes & Occ &  25.05  &  19.49  &  15.26  &  \cellcolor[HTML]{E6E6E6}19.93     &  34.56  &  28.53  &  24.41  &  \cellcolor[HTML]{E6E6E6}29.17 \\
\textbf{OccTENS-O (Ours)} & nuScenes & Occ & \textbf{27.96}   &  \textbf{21.75}  &  \textbf{16.47}  &  \cellcolor[HTML]{E6E6E6}\textbf{22.06}     &  \textbf{38.73}  &  \textbf{29.50}  &  \textbf{24.86}  &  \cellcolor[HTML]{E6E6E6}\textbf{31.03} \\

\midrule

{OccWorld-F~\cite{zheng2023occworld}} & {Waymo} & {Cam} & {17.46}  & {14.39}   &  {11.72}  &  \cellcolor[HTML]{E6E6E6}{14.52}     &     {20.94}   &  {17.33}  &  {12.27}  &  \cellcolor[HTML]{E6E6E6}{17.18}    \\
{\textbf{OccTENS-F}} & {Waymo} & {Cam} &  {\textbf{25.64}}  &  {\textbf{22.16}}  &  {\textbf{17.73}}  &   \cellcolor[HTML]{E6E6E6}{\textbf{21.84}}         &  {\textbf{29.63}}  &  {\textbf{25.40}}  &  {\textbf{23.52}} &   \cellcolor[HTML]{E6E6E6}{\textbf{26.18}}  \\
\midrule
{OccWorld-O~\cite{zheng2023occworld}} & {Waymo} & {Occ} &  {32.04}  & {25.77}   &  {23.76}  &  \cellcolor[HTML]{E6E6E6}{27.19}     &     {36.04}   &  {30.48}  &  {27.96}  &  \cellcolor[HTML]{E6E6E6}{31.49}    \\
{\textbf{OccTENS-O}} & {Waymo} & {Occ} &  {\textbf{34.17}}  &  {\textbf{27.04}}  &  {\textbf{25.47}}  &   \cellcolor[HTML]{E6E6E6}{\textbf{28.89}}         &  {\textbf{38.72}}  &  {\textbf{31.85}}  &  {\textbf{30.28}} &   \cellcolor[HTML]{E6E6E6}{\textbf{33.62}}  \\

\bottomrule
% \multicolumn{10}{l}{$\bullet$ The "-O" represents the results utilizing ground truth occupancy as input.} \\
% \multicolumn{10}{l}{$\bullet$ The "-F" represents that the input is multi-view camera images.}

\end{tabular}
% }
\caption{{\textbf{Quantitative results of 4D occupancy forecasting on nuScenes and Waymo.}} 
The ``-O'' represents the results utilizing ground truth occupancy as input.
The ``-F'' represents that the input is multi-view camera images and we use FBOCC \cite{li2023fb} to predict the occupancy from images.}
\vspace{-3mm}
\label{tab:nuscenes_main}
\end{table*}

\section{Experiments}

\subsection{Experimental Setup}
\textbf{Datasets and Metrics.} 
{We evaluate our method on the nuScenes \cite{caesar2020nuscenes} dataset and Waymo dataset \cite{sun2020scalability}. We employ the occupancy annotation in Occ3D \cite{tian2024occ3d} based on nuScenes and Waymo.}
Following common practices, we utilize a 2-second historical context (4 frames) and forecast the subsequent 3-second scenes (6 frames) unless specified. 

\noindent \textbf{Implementation Details. }
Our training period consists of 2 stages: tokenization and generation. For tokenization, we downsample the occupancy with a factor of 8. The codebook comprises 4096 nodes, and the channel dimension of the codebook entry is 128. We utilize 6 scales with [1,5,10,15,20,25] for multi-scale settings. In the tokenizer loss function, the ${\lambda}_1$, ${\lambda}_2$, ${\lambda}_3$, ${\lambda}_4$ are 10.0, 1.0, 0.3, 0.5 respectively.
For generation, we utilize 4 layers each for three blocks of our methods. The hidden dimension and head number are 128 and 4, respectively. The ${\beta}_1$ and ${\beta}_2$ are 1.0 and 1.0 respectively.

\begin{figure*}[!ht]
    \centering
    \includegraphics[width=0.93\textwidth]{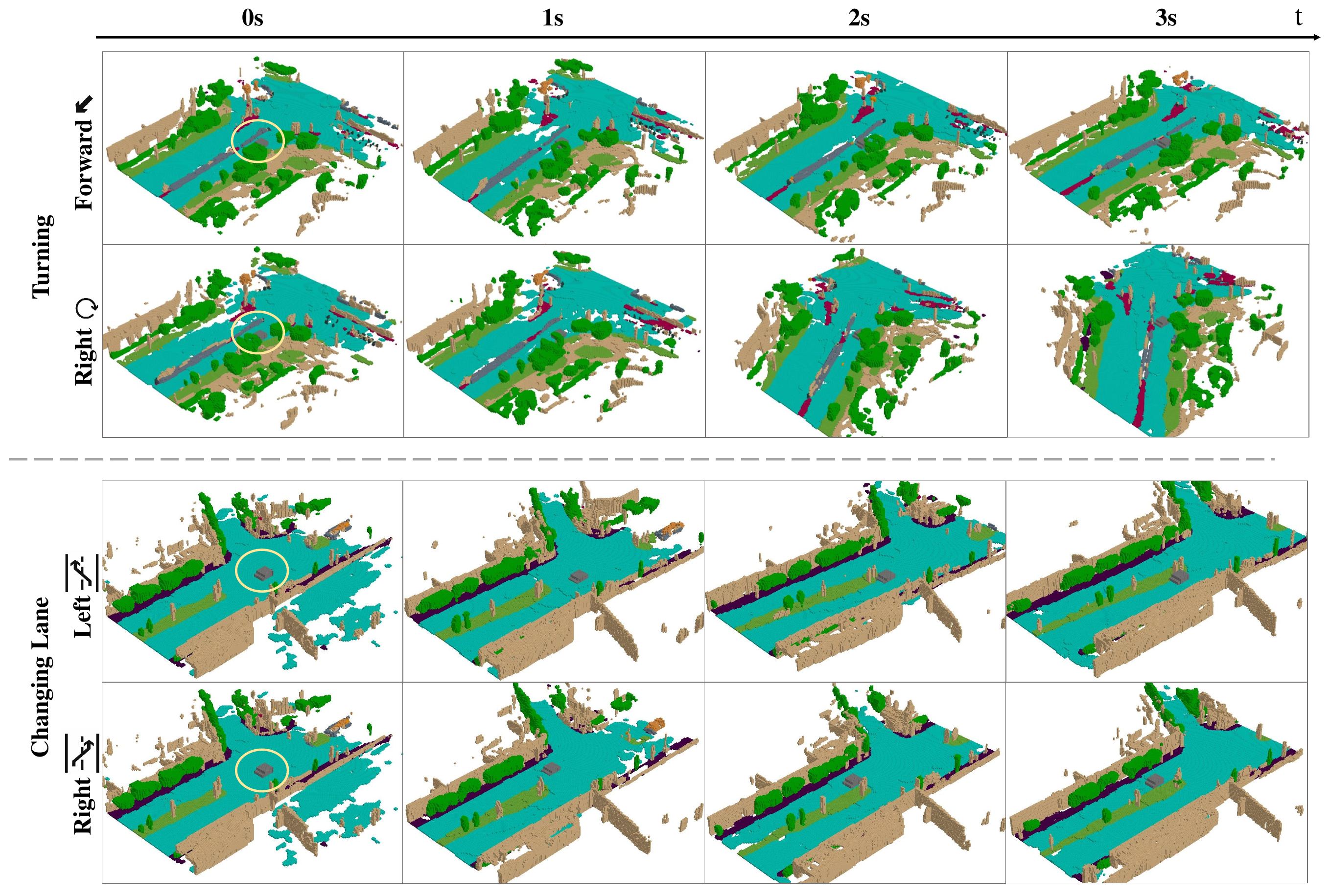}
    \caption{\textbf{Qualitative results of pose controllability}. OccTENS successfully generates results aligned with the pose input like turning (top) or changing lane (bottom), indicating the superior controllability of OccTENS.}
    \vspace{-2mm}
    \label{fig:control}
\end{figure*}

% TODO: add to supp
% \begin{table}[!t]
% \centering
% \setlength{\tabcolsep}{4pt}
% \renewcommand\arraystretch{1.1}

% \begin{tabular}{c|cc|cc}
% \hline
% Method  & \multicolumn{2}{c|}{Setting}  & \multicolumn{2}{c}{Reconstruction} \\
%         & Res. & Dim. & MIOU(\%)$\uparrow$   & IOU(\%)$\uparrow$   \\ \hline
% Occsora \cite{wang2024occsora} & 25 & 256 & 27.40 & 37.00 \\
% Occworld \cite{zheng2023occworld} & 50         & 128       &        66.38          &       62.29          \\ 
% OccLLaMA \cite{wei2024occllama} & 50         & 256       &        70.94         &        61.03         \\
% \midrule

% Ours & 25         & 128        &        \textbf{57.83}          &       \textbf{49.76} \\
% Ours & 50         & 128        &        \textbf{75.09}          & \textbf{68.96} \\
% \hline
% \end{tabular}
% \caption{Ablation study of tokenizer setting. Our multi-scale tokenizer achieves better reconstruction results.}
% \vspace{-2mm}
% \label{tab:tokenizer_ablation}
% \end{table}

\subsection{Main Results}
{The evaluation of OccTENS includes two tasks: 4D occupancy forecasting task and the motion planning task. The 4D occupancy forecasting task aims to forecast the future observation of 3D occupancy scenes. The motion planning task plans a trajectory for the ego vehicle.}

% TODO: add to supp
% \noindent \textbf{3D Occupancy Reconstruction}
% We compare the reconstruction performance of different hyperparameters of the tokenizer, results shown in Tab.~\ref{tab:tokenizer_ablation}. We also show the reconstruction performance of the existing methods for comparison. We observe that our multi-scale scene tokenizer outperforms the baseline by a large margin, demonstrating the superiority of multi-scale quantization. 
% We also observe that a larger resolution of the latent BEV feature map delivers better performance. However, the increase in resolution will also lead to a significant increase in the number of tokens after discretization, which influences the performance in the generation stage. 
% More results are shown in supplementary materials.

\textbf{4D Occupancy Forecasting}
In this experiment, we compare our method with state-of-the-art approaches on the 4D occupancy forecasting task. {In fairness, we only report the occupancy generation with \textbf{only} historical observation as input, with no additional auxiliary inputs.} Following common practice, we conduct our evaluation in two highlights: (1) using ground-truth 3D occupancy data (-O); and (2) using predicted results from FBOCC~\cite{li2023fb} based on camera data (-F). The results are shown in Tab.~\ref{tab:nuscenes_main}.
We observe that our OccTENS achieves significant performance gain over existing methods in short-time forecasting within 3 seconds. 
These results highlight the strong predictive performance of OccTENS, which sets the state-of-the-art on the nuScenes dataset and Waymo dataset. Moreover, as shown in Fig.~\ref{fig:long_generation}, OccTENS demonstrates significantly improved performance in long-term sequence generation, producing coherent, high-fidelity occupancy scenes that closely mirror 3D dynamics.

\begin{table}[!t]
\centering
% \setlength{\tabcolsep}{1.4mm}
% \renewcommand\arraystretch{1.2}
% \vspace{-3mm}

\setlength{\tabcolsep}{0.01\linewidth}
\resizebox{1.0\linewidth}{!}{
\begin{tabular}{l|cccc|cccc}
% \cline{1-11}
\toprule
\multirow{2}{*}{Method} & \multicolumn{4}{c|}{L2(m)$\downarrow$} & \multicolumn{4}{c}{Coll.(\%)$\downarrow$} \\
& 1s & 2s & 3s & \cellcolor[HTML]{E6E6E6}Avg. & 1s & 2s & 3s & \cellcolor[HTML]{E6E6E6}Avg. \\ 
% \cline{1-11}
\midrule
OccWorld \cite{zheng2023occworld} & 0.43 & 1.08 & 1.99 & \cellcolor[HTML]{E6E6E6}1.17 & 0.07 & 0.38 & 1.35 & \cellcolor[HTML]{E6E6E6}0.60 \\
OccLlama \cite{wei2024occllama}& 0.37 & 1.02 & 2.03 &  \cellcolor[HTML]{E6E6E6}1.14 & 0.04 & 0.24 & 1.20 & \cellcolor[HTML]{E6E6E6}0.49 \\

\textbf{Ours} &  0.39 & \textbf{1.02}  & \textbf{1.96} & \cellcolor[HTML]{E6E6E6}\textbf{1.12} & 0.08 & 0.25 & \textbf{1.12} & \cellcolor[HTML]{E6E6E6}\textbf{0.48} \\

\midrule
UniAD \cite{hu2023planning} & 0.48 & 0.96 & 1.65 &  \cellcolor[HTML]{E6E6E6}1.03 & 0.05 & 0.17 & 0.71 & \cellcolor[HTML]{E6E6E6}0.31 \\

VAD \cite{jiang2023vad} & 0.41 & 0.70 &  1.05 &  \cellcolor[HTML]{E6E6E6}0.72 & 0.07 &  0.17 &  0.41 & \cellcolor[HTML]{E6E6E6}0.22 \\

PPAD \cite{chen2024ppad} & 0.31 & 0.56 & 0.87 &  \cellcolor[HTML]{E6E6E6} 0.58 & 0.08 &  0.12 & 0.38 & \cellcolor[HTML]{E6E6E6}0.19 \\

GenAD \cite{zheng2024genad} & 0.28 & 0.49 & 0.78 &  \cellcolor[HTML]{E6E6E6}0.52 &  0.08 & 0.14 & 0.34 & \cellcolor[HTML]{E6E6E6}0.19 \\

LAW \cite{li2024enhancing} & 0.24 & 0.46 & 0.76 &  \cellcolor[HTML]{E6E6E6}0.49 & 0.08 & 0.10 & 0.39 & \cellcolor[HTML]{E6E6E6}0.19 \\
\bottomrule

\end{tabular}
}

\caption{\textbf{Motion planning results.} We can see that OccTENS gets a better planning performance. We include SOTA end-to-end planning methods for reference.}
\label{tab:planning}
\vspace{-2mm}
\end{table}

\begin{table}[!t]
\centering
\setlength{\tabcolsep}{2mm}
\renewcommand\arraystretch{0.9}
% \vspace{-1mm}

\resizebox{\linewidth}{!}{
\begin{tabular}{l|ccc}
% \cline{1-11}
\toprule
\multirow{2}{*}{Scale Num.}  & \multicolumn{3}{c}{Avg. Gen.} \\
 & mIoU(\%) & IoU(\%) & Latency(s) \\
% \cline{1-11}
\midrule
OccWorld & 17.14 & 26.63 & 0.35 \\
OccLlama & 19.93 & 29.17 & - \\
OccSora & - & - & \~20 \\
\midrule

Ours-2-scales   & 18.35 & 27.31 & \textbf{0.21} \\
% Ours-3 & 70.14 & 65.53 & 19.24 & 27.32 & 0.32 \\
Ours-4-scales  & 21.24 & 29.72 & 0.34 \\
\cellcolor[HTML]{E6E6E6}Ours-6-scales & \cellcolor[HTML]{E6E6E6}22.06 & \cellcolor[HTML]{E6E6E6}31.03 & \cellcolor[HTML]{E6E6E6}0.56 \\
Ours-8-scales  & \textbf{22.23} & \textbf{31.24} & 0.93 \\
\bottomrule

% \multicolumn{11}{l}{The best and second-best performances are represented by \textbf{bold} and \underline{underline}
\end{tabular}
}
% \vspace{-6mm}
\caption{\textbf{Efficiency analysis}. OccTENS can get a higher performance with a faster inference time. The ``-'' represents the unreported results. The gray square represents the setting we used in the previous experiment.}
\vspace{-4mm}
\label{tab:efficiency}
\end{table}

\textbf{Motion Planning}
We compare the motion planning performance of OccTENS with several strong baselines that utilize the same inputs and supervision methods. The results are shown in Tab.~\ref{tab:planning}. We observe that overall OccTENS surpasses existing occupancy world models in the motion planning task. Interestingly, {OccTENS achieves improvements in planning results, especially at 2s and 3s}. We attribute this to the fact that OccTENS explicitly models the dynamics during occupancy generation, enabling robust long-horizon trajectory optimization through accurate anticipation of future occupancy changes, which is critical for navigating complex, unstructured scenarios. We also include the results of several SOTA planning methods for reference. While our results are lower than methods like UniAD, this is expected since UniAD benefits from extensive multi-task supervision (e.g., detection, tracking, and map annotation). In contrast, OccTENS focuses on a more self-contained and generative world modeling setup, relying solely on occupancy-based representations without such auxiliary labels. This comparison highlights the complementary nature of these approaches and underscores the potential of occupancy-centric modeling even under reduced supervision.

% \begin{table}[!t]
% \centering
% \setlength{\tabcolsep}{4pt}
% % 这里算一下flops，比较完整的时间，各个部分的latency
% % 这里可以吧para和latency分开，上下两部分
% \begin{tabular}{c|ccc}
% \toprule
% Method   & Latency(s)   & MIOU(\%)   & IOU(\%)   \\
% \midrule
% OccSora \cite{wang2024occsora}                 & $\sim$ 100 &        -          &       -          \\ 
% OccWorld \cite{zheng2023occworld}                & 0.35 &        17.14          &       26.63          \\ 
% OccLLaMA \cite{wei2024occllama}                & - &        19.93          &       29.17         \\ 
% \midrule
% Ours (NTP)                 & $\sim$ 5 &        19.20          &  27.32               \\ 

% Ours                & 0.56 &         22.06          &  31.03               \\ 

% \bottomrule
% \end{tabular}
% \caption{\textbf{Efficiency analysis.} The latency refers to the inference time of generating a scene with 6 frames. The NTP represents that we utilize next token prediction during generation.}
% % \vspace{-2mm}
% \label{tab:efficiency}
% \end{table}

\begin{table}[!t]
\centering

\begin{tabular}{ccc|ccc}
\toprule
{Decomp} & {TNSP} & {Pose Agg}   & {Latency(s)}   & {mIoU(\%)}   & {IoU(\%)}   \\
\midrule
 & {\checkmark} & {\checkmark}               & {2.47} &        {17.14}          &       {26.63}          \\ 
{\checkmark} &  & {\checkmark}                 & {5.13} &        {19.20}          &  {27.32}               \\ 
{\checkmark} & {\checkmark} &                 & {0.59} &        {19.93}          &       {29.17}         \\ 

{\checkmark} & {\checkmark} & {\checkmark}                & {\textbf{0.56}} &         {\textbf{22.06}}          &  {\textbf{31.03}}               \\ 

\bottomrule
\end{tabular}
\caption{{\textbf{Component-wise ablation.} Each component contributes to the
improvement of the performance.}}
\label{tab:element_ablation}
\vspace{-5mm}
\end{table}

\subsection{Ablation Results}
To delve into the effect of each module, we conduct a comprehensive ablation study on OccTENS on efficiency and pose controllability.

% TODO: add to supp
% \noindent \textbf{Long-term Generation.} To evaluate the long-term generation capabilities of OccTENS, we conducted a series of experiments comparing its performance against OccWorld \cite{zheng2023occworld}. As shown in Fig.~\ref{fig:long_generation}, we observe that OccWorld exhibits repetition artifacts when generating long time series. Specifically, after a certain number of time steps, the model begins to produce repetitive patterns, which diminishes the fidelity of the generated 3D scenes.

% In contrast, OccTENS demonstrates significantly improved performance in long-term sequence generation, producing coherent, high-fidelity occupancy scenes that closely mirror real-world 3D dynamics. For example, the geometry of the bus in the first sequence is well-maintained over time and our model avoids the common pitfalls of next-token autoregressive models. 

\noindent \textbf{Efficiency.}
The latency is of great significance for the deployment of the autonomous driving system. In this experiment, we compare OccTENS with existing works. 
% As far as we know, OccLLaMA \cite{wei2024occllama} is not publicly available and does not report their inference time. Moreover, OccWorld is not a standard AR model that utilizes next-token prediction. Thus we design an AR model by adapting our method for next-token prediction. Specifically, we utilize the tokenizer in OccWorld \cite{zheng2023occworld} while employing the same generation architecture as ours, denoted as ``Ours (NTP)''. The results are shown in Tab.~\ref{tab:efficiency}. We can see that compared with next-token prediction (Ours NTP), the next-scale prediction has a much lower latency while demonstrating better performance. This indicates the superiority of the next-scale prediction rather than next-token prediction in 4D occupancy world model.
It should be emphasized that OccTENS can indeed achieve competitive speed if we reduce the number of scales. Our choice to use 6 scales aims to strike an optimal trade-off between performance and efficiency. However, we can observe that the OccTENS with 2 scales outperforms OccWorld in both generation performance and inference time, demonstrating the effectiveness and efficiency of the proposed architecture. 

Our findings show that increasing the number of scales consistently improves both reconstruction and generation performance. However, this improvement comes at the cost of increased latency, indicating that a trade-off between performance and efficiency must be maintained.

% It is worth mentioning that OccWorld has a very fast inference speed, because it utilizes position-wise temporal attention, where the tokens of one frame are generated in a single forward step. However, we can observe that the OccTENS with 2 scales outperforms OccWorld in both generation performance and inference time, demonstrating the effectiveness and efficiency of the proposed architecture. 

\noindent \textbf{Pose Controllability}
Controllability refers to the model’s capacity to precisely adhere to these inputs, ensuring that the generated scenes reflect the specified conditions with high fidelity. Pose control is particularly critical, as it ensures that the model generates scenes from the correct viewpoint and perspective. We manipulate the camera pose inputs and measure the generation results in terms of alignment. As shown in Fig.~\ref{fig:control}, OccTENS can generate the corresponding results that are collaboratively aligned with the conditional motion input, indicating the powerful generalization ability of our method.
% \vspace{-1.5mm}

\noindent {\textbf{Component-wise ablation} We conduct a component-wise ablation study to illustrate the effectiveness of our design, including:
(1) decomposing Architecture (with/without),
(2) next-token prediction (NTP) vs. temporal next-scale prediction (TNSP),
(3) Multi-modal pose aggregation (with/without).
The results are shown in Table~\ref{tab:element_ablation}. We can observe that each component of our architecture contributes to the improvement of the performance. We can also observe that our design of decomposing architecture (first row) and temporal next-scale prediction (second row) can reduce the inference latencies of the occupancy world model.}

%% file: sections/5_conclusion.tex
\section{Conclusions}
In this paper, we present OccTENS, an autoregressive occupancy world model designed to generate controllable, high-fidelity long-term occupancy generation while maintaining computational efficiency. By reformulating the occupancy world model as a temporal next-scale prediction (TENS) task, OccTENS is capable of effectively managing the temporal causality and spatial relationships of occupancy sequences. Extensive evaluations demonstrate the superiority of OccTENS on fidelity, efficiency, and controllability, surpassing existing methods. These results highlight the potential of OccTENS to facilitate real-time applications in autonomous driving, paving the way for future advancements in world models.

%% file: main.bib
@String(CVPR= {IEEE Conf. Comput. Vis. Pattern Recog.})

@String(ICCV= {Int. Conf. Comput. Vis.})

@String(ECCV= {Eur. Conf. Comput. Vis.})

@String(CVPR  = {CVPR})

@String(ICCV  = {ICCV})

@String(ECCV  = {ECCV})

@article{zheng2023occworld,
  title={Occworld: Learning a 3d occupancy world model for autonomous driving},
  author={Zheng, Wenzhao and Chen, Weiliang and Huang, Yuanhui and Zhang, Borui and Duan, Yueqi and Lu, Jiwen},
  journal={ECCV},
  year={2024}
}

@inproceedings{wei2023surroundocc,
  title={Surroundocc: Multi-camera 3d occupancy prediction for autonomous driving},
  author={Wei, Yi and Zhao, Linqing and Zheng, Wenzhao and Zhu, Zheng and Zhou, Jie and Lu, Jiwen},
  booktitle={ICCV},
  year={2023}
}

@article{tian2024visual,
  title={Visual autoregressive modeling: Scalable image generation via next-scale prediction},
  author={Tian, Keyu and Jiang, Yi and Yuan, Zehuan and Peng, Bingyue and Wang, Liwei},
  journal={Neurips},
  year={2024}
}

@inproceedings{berman2018lovasz,
  title={The lov{\'a}sz-softmax loss: A tractable surrogate for the optimization of the intersection-over-union measure in neural networks},
  author={Berman, Maxim and Triki, Amal Rannen and Blaschko, Matthew B},
  booktitle={CVPR},
  year={2018}
}

@inproceedings{cao2022monoscene,
  title={Monoscene: Monocular 3d semantic scene completion},
  author={Cao, Anh-Quan and De Charette, Raoul},
  booktitle={CVPR},
  year={2022}
}

@article{gao2024vista,
  title={Vista: A Generalizable Driving World Model with High Fidelity and Versatile Controllability},
  author={Gao, Shenyuan and Yang, Jiazhi and Chen, Li and Chitta, Kashyap and Qiu, Yihang and Geiger, Andreas and Zhang, Jun and Li, Hongyang},
  journal={Neurips},
  year={2024}
}

@inproceedings{caesar2020nuscenes,
  title={nuscenes: A multimodal dataset for autonomous driving},
  author={Caesar, Holger and Bankiti, Varun and Lang, Alex H and Vora, Sourabh and Liong, Venice Erin and Xu, Qiang and Krishnan, Anush and Pan, Yu and Baldan, Giancarlo and Beijbom, Oscar},
  booktitle={CVPR},
  year={2020}
}

@article{wang2023drivedreamer,
  title={DriveDreamer: Towards Real-world-driven World Models for Autonomous Driving},
  author={Wang, Xiaofeng and Zhu, Zheng and Huang, Guan and Chen, Xinze and Zhu, Jiagang and Lu, Jiwen},
  journal={ECCV},
  year={2024}
}

@article{lu2023wovogen,
  title={Wovogen: World volume-aware diffusion for controllable multi-camera driving scene generation},
  author={Lu, Jiachen and Huang, Ze and Zhang, Jiahui and Yang, Zeyu and Zhang, Li},
  journal={ECCV},
  year={2024}
}

@inproceedings{wang2024driving,
  title={Driving into the future: Multiview visual forecasting and planning with world model for autonomous driving},
  author={Wang, Yuqi and He, Jiawei and Fan, Lue and Li, Hongxin and Chen, Yuntao and Zhang, Zhaoxiang},
  booktitle={CVPR},
  year={2024}
}

@article{zheng2024genad,
  title={Genad: Generative end-to-end autonomous driving},
  author={Zheng, Wenzhao and Song, Ruiqi and Guo, Xianda and Chen, Long},
  journal={ECCV},
  year={2024}
}

@article{jiang2024dive,
  title={DiVE: DiT-based Video Generation with Enhanced Control},
  author={Jiang, Junpeng and Hong, Gangyi and Zhou, Lijun and Ma, Enhui and Hu, Hengtong and Zhou, Xia and Xiang, Jie and Liu, Fan and Yu, Kaicheng and Sun, Haiyang and others},
  journal={arXiv preprint arXiv:2409.01595},
  year={2024}
}

@article{wei2024occllama,
  title={OccLLaMA: An Occupancy-Language-Action Generative World Model for Autonomous Driving},
  author={Wei, Julong and Yuan, Shanshuai and Li, Pengfei and Hu, Qingda and Gan, Zhongxue and Ding, Wenchao},
  journal={arXiv preprint arXiv:2409.03272},
  year={2024}
}

@article{tian2024occ3d,
  title={Occ3d: A large-scale 3d occupancy prediction benchmark for autonomous driving},
  author={Tian, Xiaoyu and Jiang, Tao and Yun, Longfei and Mao, Yucheng and Yang, Huitong and Wang, Yue and Wang, Yilun and Zhao, Hang},
  journal={Neurips},
  volume={36},
  year={2024}
}

@article{li2023fb,
  title={Fb-occ: 3d occupancy prediction based on forward-backward view transformation},
  author={Li, Zhiqi and Yu, Zhiding and Austin, David and Fang, Mingsheng and Lan, Shiyi and Kautz, Jan and Alvarez, Jose M},
  journal={arXiv preprint arXiv:2307.01492},
  year={2023}
}

@inproceedings{hu2023planning,
  title={Planning-oriented autonomous driving},
  author={Hu, Yihan and Yang, Jiazhi and Chen, Li and Li, Keyu and Sima, Chonghao and Zhu, Xizhou and Chai, Siqi and Du, Senyao and Lin, Tianwei and Wang, Wenhai and others},
  booktitle={CVPR},
  year={2023}
}

@inproceedings{jiang2023vad,
  title={Vad: Vectorized scene representation for efficient autonomous driving},
  author={Jiang, Bo and Chen, Shaoyu and Xu, Qing and Liao, Bencheng and Chen, Jiajie and Zhou, Helong and Zhang, Qian and Liu, Wenyu and Huang, Chang and Wang, Xinggang},
  booktitle={ICCV},
  year={2023}
}

@inproceedings{hu2022st,
  title={St-p3: End-to-end vision-based autonomous driving via spatial-temporal feature learning},
  author={Hu, Shengchao and Chen, Li and Wu, Penghao and Li, Hongyang and Yan, Junchi and Tao, Dacheng},
  booktitle={ECCV},
  year={2022},
}

@article{huang2023differentiable,
  title={Differentiable integrated motion prediction and planning with learnable cost function for autonomous driving},
  author={Huang, Zhiyu and Liu, Haochen and Wu, Jingda and Lv, Chen},
  journal={IEEE TNNLS},
  year={2023},
  publisher={IEEE}
}

@inproceedings{yang2024unipad,
  title={Unipad: A universal pre-training paradigm for autonomous driving},
  author={Yang, Honghui and Zhang, Sha and Huang, Di and Wu, Xiaoyang and Zhu, Haoyi and He, Tong and Tang, Shixiang and Zhao, Hengshuang and Qiu, Qibo and Lin, Binbin and others},
  booktitle={CVPR},
  year={2024}
}

@article{wang2024occsora,
  title={OccSora: 4D Occupancy Generation Models as World Simulators for Autonomous Driving},
  author={Wang, Lening and Zheng, Wenzhao and Ren, Yilong and Jiang, Han and Cui, Zhiyong and Yu, Haiyang and Lu, Jiwen},
  journal={arXiv preprint arXiv:2405.20337},
  year={2024}
}

@inproceedings{sun2020scalability,
  title={Scalability in perception for autonomous driving: Waymo open dataset},
  author={Sun, Pei and Kretzschmar, Henrik and Dotiwalla, Xerxes and Chouard, Aurelien and Patnaik, Vijaysai and Tsui, Paul and Guo, James and Zhou, Yin and Chai, Yuning and Caine, Benjamin and others},
  booktitle={CVPR},
  year={2020}
}

@inproceedings{agro2024uno,
  title={Uno: Unsupervised occupancy fields for perception and forecasting},
  author={Agro, Ben and Sykora, Quinlan and Casas, Sergio and Gilles, Thomas and Urtasun, Raquel},
  booktitle={CVPR},
  year={2024}
}

@inproceedings{diehl2025dio,
  title={DIO: Decomposable Implicit 4D Occupancy-Flow World Model},
  author={Diehl, Christopher and Sykora, Quinlan and Agro, Ben and Gilles, Thomas and Casas, Sergio and Urtasun, Raquel},
  booktitle={CVPR},
  year={2025}
}

@inproceedings{tong2023scene,
  title={Scene as occupancy},
  author={Tong, Wenwen and Sima, Chonghao and Wang, Tai and Chen, Li and Wu, Silei and Deng, Hanming and Gu, Yi and Lu, Lewei and Luo, Ping and Lin, Dahua and others},
  booktitle={ICCV},
  year={2023}
}

@article{ha2018world,
  title={World models},
  author={Ha, David and Schmidhuber, J{\"u}rgen},
  journal={arXiv preprint arXiv:1803.10122},
  year={2018}
}

@article{hu2023gaia,
  title={Gaia-1: A generative world model for autonomous driving},
  author={Hu, Anthony and Russell, Lloyd and Yeo, Hudson and Murez, Zak and Fedoseev, George and Kendall, Alex and Shotton, Jamie and Corrado, Gianluca},
  journal={arXiv preprint arXiv:2309.17080},
  year={2023}
}

@article{zhao2024drivedreamer,
  title={Drivedreamer-2: Llm-enhanced world models for diverse driving video generation},
  author={Zhao, Guosheng and Wang, Xiaofeng and Zhu, Zheng and Chen, Xinze and Huang, Guan and Bao, Xiaoyi and Wang, Xingang},
  journal={arXiv preprint arXiv:2403.06845},
  year={2024}
}

@article{su2024text2street,
  title={Text2Street: Controllable Text-to-image Generation for Street Views},
  author={Su, Jinming and Gu, Songen and Duan, Yiting and Chen, Xingyue and Luo, Junfeng},
  journal={arXiv preprint arXiv:2402.04504},
  year={2024}
}

@article{zhang2023learning,
  title={Learning unsupervised world models for autonomous driving via discrete diffusion},
  author={Zhang, Lunjun and Xiong, Yuwen and Yang, Ze and Casas, Sergio and Hu, Rui and Urtasun, Raquel},
  journal={arXiv preprint arXiv:2311.01017},
  year={2023}
}

@article{zyrianov2024lidardm,
  title={LidarDM: Generative LiDAR Simulation in a Generated World},
  author={Zyrianov, Vlas and Che, Henry and Liu, Zhijian and Wang, Shenlong},
  journal={arXiv preprint arXiv:2404.02903},
  year={2024}
}

@article{huang2023tri,
    title={Tri-Perspective View for Vision-Based 3D Semantic Occupancy Prediction},
    author={Huang, Yuanhui and Zheng, Wenzhao and Zhang, Yunpeng and Zhou, Jie and Lu, Jiwen },
    journal={arXiv preprint arXiv:2302.07817},
    year={2023}
}

@inproceedings{ma2024cotr,
  title={Cotr: Compact occupancy transformer for vision-based 3d occupancy prediction},
  author={Ma, Qihang and Tan, Xin and Qu, Yanyun and Ma, Lizhuang and Zhang, Zhizhong and Xie, Yuan},
  booktitle={CVPR},
  year={2024}
}

@inproceedings{wang2024panoocc,
  title={Panoocc: Unified occupancy representation for camera-based 3d panoptic segmentation},
  author={Wang, Yuqi and Chen, Yuntao and Liao, Xingyu and Fan, Lue and Zhang, Zhaoxiang},
  booktitle={CVPR},
  year={2024}
}

@article{zheng2024monoocc,
  title={Monoocc: Digging into monocular semantic occupancy prediction},
  author={Zheng, Yupeng and Li, Xiang and Li, Pengfei and Zheng, Yuhang and Jin, Bu and Zhong, Chengliang and Long, Xiaoxiao and Zhao, Hao and Zhang, Qichao},
  journal={arXiv preprint arXiv:2403.08766},
  year={2024}
}

@inproceedings{pan2024renderocc,
  title={Renderocc: Vision-centric 3d occupancy prediction with 2d rendering supervision},
  author={Pan, Mingjie and Liu, Jiaming and Zhang, Renrui and Huang, Peixiang and Li, Xiaoqi and Xie, Hongwei and Wang, Bing and Liu, Li and Zhang, Shanghang},
  booktitle={ICRA},
  year={2024},
}

@inproceedings{huang2024selfocc,
  title={Selfocc: Self-supervised vision-based 3d occupancy prediction},
  author={Huang, Yuanhui and Zheng, Wenzhao and Zhang, Borui and Zhou, Jie and Lu, Jiwen},
  booktitle={CVPR},
  year={2024}
}

@article{li2024uniscene,
  title={UniScene: Unified Occupancy-centric Driving Scene Generation},
  author={Li, Bohan and Guo, Jiazhe and Liu, Hongsi and Zou, Yingshuang and Ding, Yikang and Chen, Xiwu and Zhu, Hu and Tan, Feiyang and Zhang, Chi and Wang, Tiancai and others},
  journal={arXiv preprint arXiv:2412.05435},
  year={2024}
}

@article{gu2024dome,
  title={Dome: Taming diffusion model into high-fidelity controllable occupancy world model},
  author={Gu, Songen and Yin, Wei and Jin, Bu and Guo, Xiaoyang and Wang, Junming and Li, Haodong and Zhang, Qian and Long, Xiaoxiao},
  journal={arXiv preprint arXiv:2410.10429},
  year={2024}
}

@inproceedings{chen2024ppad,
  title={Ppad: Iterative interactions of prediction and planning for end-to-end autonomous driving},
  author={Chen, Zhili and Ye, Maosheng and Xu, Shuangjie and Cao, Tongyi and Chen, Qifeng},
  booktitle={ECCV},
  pages={239--256},
  year={2024},
  organization={Springer}
}

@article{li2024enhancing,
  title={Enhancing end-to-end autonomous driving with latent world model},
  author={Li, Yingyan and Fan, Lue and He, Jiawei and Wang, Yuqi and Chen, Yuntao and Zhang, Zhaoxiang and Tan, Tieniu},
  journal={arXiv preprint arXiv:2406.08481},
  year={2024}
}

@article{xu2025occ,
  title={Occ-LLM: Enhancing Autonomous Driving with Occupancy-Based Large Language Models},
  author={Xu, Tianshuo and Lu, Hao and Yan, Xu and Cai, Yingjie and Liu, Bingbing and Chen, Yingcong},
  journal={arXiv preprint arXiv:2502.06419},
  year={2025}
}

@inproceedings{wang2024drivedreamer,
  title={DriveDreamer: Towards Real-World-Drive World Models for Autonomous Driving},
  author={Wang, Xiaofeng and Zhu, Zheng and Huang, Guan and Chen, Xinze and Zhu, Jiagang and Lu, Jiwen},
  booktitle={ECCV},
  year={2024},
}

@article{gao2023magicdrive,
  title={Magicdrive: Street view generation with diverse 3d geometry control},
  author={Gao, Ruiyuan and Chen, Kai and Xie, Enze and Hong, Lanqing and Li, Zhenguo and Yeung, Dit-Yan and Xu, Qiang},
  journal={arXiv preprint arXiv:2310.02601},
  year={2023}
}

@article{li2024autoregressive,
  title={Autoregressive image generation without vector quantization},
  author={Li, Tianhong and Tian, Yonglong and Li, He and Deng, Mingyang and He, Kaiming},
  journal={Neurips},
  year={2024}
}

@article{deng2024autoregressive,
  title={Autoregressive Video Generation without Vector Quantization},
  author={Deng, Haoge and Pan, Ting and Diao, Haiwen and Luo, Zhengxiong and Cui, Yufeng and Lu, Huchuan and Shan, Shiguang and Qi, Yonggang and Wang, Xinlong},
  journal={arXiv preprint arXiv:2412.14169},
  year={2024}
}

@article{swerdlow2024street,
  title={Street-view image generation from a bird's-eye view layout},
  author={Swerdlow, Alexander and Xu, Runsheng and Zhou, Bolei},
  journal={IEEE RAL},
  year={2024},
  publisher={IEEE}
}

@article{yang2023bevcontrol,
  title={Bevcontrol: Accurately controlling street-view elements with multi-perspective consistency via bev sketch layout},
  author={Yang, Kairui and Ma, Enhui and Peng, Jibin and Guo, Qing and Lin, Di and Yu, Kaicheng},
  journal={arXiv preprint arXiv:2308.01661},
  year={2023}
}
